\documentclass{article}

\usepackage{amsmath,amsfonts,amssymb,times,graphicx,natbib,algorithm,algorithmic,hyperref}

\usepackage{subcaption}
\usepackage{algorithm}
\usepackage{algorithmic}
\usepackage{amsmath}
\usepackage{amssymb}
\usepackage{mathtools}
\usepackage{pgf}
\usepackage{import}

\usepackage{microtype}      
\usepackage{xfrac}

\newcommand\numberthis{\addtocounter{equation}{1}\tag{\theequation}}

\DeclareMathAlphabet{\mathcal}{OMS}{cmsy}{m}{n}

\DeclareMathOperator*{\argmin}{argmin}

\DeclarePairedDelimiterX{\inner}[2]{\langle}{\rangle}{#1, #2}

\newcommand{\xn}{\textbf{x}^{(n)}}

\newcommand{\obj}{J}

\usepackage[accepted]{whi2018}

\icmltitlerunning{Techniques for visualizing LSTMs applied to electrocardiograms}

\begin{document}

\graphicspath{{figures/}}
\makeatletter
\def\input@path{{figures/}}
\makeatother

\twocolumn[
\icmltitle{Techniques for visualizing LSTMs applied to electrocardiograms}


\icmlsetsymbol{equal}{*}

\begin{icmlauthorlist}
	\icmlauthor{Jos van der Westhuizen}{cam}
	\icmlauthor{Joan Lasenby}{cam}
\end{icmlauthorlist}

\icmlaffiliation{cam}{Department of Engineering, Cambridge University, Cambridge, UK}

\icmlcorrespondingauthor{Jos van der Westhuizen}{jv365@cam.ac.uk}

\icmlkeywords{interpretability, transparency, LSTM, ECG}

\vskip 0.3in
]


\printAffiliationsAndNotice{}

\begin{abstract}
This paper explores four different visualization techniques for long short-term memory (LSTM) networks applied to continuous-valued time series. On the datasets analysed, we find that the best visualization technique is to learn an input deletion mask that optimally reduces the true class score. With a specific focus on single-lead electrocardiograms from the MIT-BIH arrhythmia dataset, we show that salient input features for the LSTM classifier align well with medical theory.
\end{abstract}

\section{Introduction}
Long short-term memory (LSTM) networks have been shown to be effective on medical datasets \citep{lipton_learning_2015,ChoiDoctorAIPredicting2016,RajkomarScalableaccuratedeep2018,jagannatha2016bidirectional,TeijeiroArrhythmiaClassificationAbductive2017,CliffordAFClassificationshort2017}. There is a snag, however. Researchers struggle to understand exactly how the self-adapted LSTMs do what they do. Hence, two main points drive a deeper understanding of LSTMs. First, clinicians need to be able to explain how a model's decisions are based on a patient's data. Second, models that achieve breakthrough performance may have identified patterns in the data that practitioners would like to understand.

Previous work has provided insights into the operation of the LSTM \citep{KarpathyVisualizingUnderstandingRecurrent2015,LiVisualizingUnderstandingNeural2016}, however, this was focussed on the discrete-valued sequences of natural language processing tasks. Conversely, our work explores visualization techniques for understanding LSTMs applied to continuous-valued sequences, specifically, single-lead electrocardiograms (ECGs). We learn from recently proposed visualization techniques, which we describe next.

\section{Related work}
A common form of interpretability for deep learning models is through attention mechanisms \citep{XuShowAttendTell2015b,BahdanauNeuralMachineTranslation2015,DemingGeneticArchitectDiscovering2016,RajkomarScalableaccuratedeep2018,ChingOpportunitiesObstaclesDeep2018}. By revealing which input features are used for different outputs, the attention mechanisms provide insights into the model's decision-making process. In the clinical domain, \citet{ChoiRETAINInterpretablePredictive2016a} leveraged attention mechanisms to highlight which aspects of a patient's medical history were most relevant for making diagnoses. \citet{ChoiGRAMGraphbasedAttention2017} later extended this work to take into account the structure of disease ontologies and found that the concepts represented by the model aligned with medical knowledge -- an exercise we strive to imitate. However, interpretation strategies that rely on attention mechanisms do not provide insight into the logic used by the attention layer.

Backpropagation-based methods are also popular for interpreting deep learning models. They have the signal from a target output neuron backpropagated to the input layer. The simplest form of this was proposed by \citet{SimonyanDeepConvolutionalNetworks2013}, and many proposals since have improved utility \citep{BachPixelWiseExplanationsNonLinear2015,KindermansInvestigatinginfluencenoise2016,SpringenbergStrivingSimplicityAll2014,MahendranSalientDeconvolutionalNetworks2016}. However, as with the attention mechanisms and deconvolution \citep{ZeilerVisualizingUnderstandingConvolutional2014}, many of these visualization techniques require architectural modifications \citep{FongInterpretableExplanationsBlack2017a}.

On the other hand, perturbation-based interpretation approaches do not require changes in the model architecture; instead, they change parts of the input and observe the impact on the output of the network. These include visualizing the drop in classification score as constant value masks are applied to different input patches of images \citep{ZeilerVisualizingUnderstandingConvolutional2014}. A recent study by \citet{FongInterpretableExplanationsBlack2017a} proposed to use gradients to learn the minimal input ``deletions'' that minimize the class score. This technique requires access to only the model's inputs and outputs and provides aesthetically pleasing explanations of image input salience. We modify this technique to visualize continuous inputs that are analysed by LSTMs.

Visualizations of recurrent neural networks (RNNs) have successfully been explored in natural language processing \citep{LiVisualizingUnderstandingNeural2016,KarpathyVisualizingUnderstandingRecurrent2015}.  In both these studies, the aim was to improve the understanding of RNNs by leveraging human knowledge of the structured language. Such well-understood discrete input lends itself to interpretable isolated explanations for the importance of each input to the model. In the medical paradigm, firstly, we are less certain about the underlying biological processes that govern the generation of physiological signals, and secondly, due to inputs being continuous, isolation of salient features is more difficult. RNNs have also been visualized by \citet{LanchantinDeepmotifdashboard2017} in the domain of DNA sequences. Here again, the DNA inputs are discrete compared with the continuous-valued medical time series that we analyse.

\section{Visualization techniques}
Before describing the four visualization techniques explored, some preliminaries are required; each input sequence $ \textbf{x}_{1:T} $ with $ T $ time steps has a corresponding class label $ c $. An LSTM provides pre-softmax activations $ \textbf{s}\in\mathbb{R}^{C} $ for each input sequence, with $ s_c $ denoting the score for the correct class $ c $ of the input sequence, and $ C $ the number of classes.

\subsection{Temporal output score}
A first approach to understanding LSTM classifications is to illustrate the progression of model decisions over time for a specific input sequence; i.e., incrementally longer subsequences of the original time series are classified and visualized. More formally, for a sequence $ \textbf{x}_{1:T} $ with length $ T $ we classify the prefix\footnote{We use the term prefix to refer to a subsequence at the start of the original sequence.} $ \textbf{x}_{1:t} $ of the original sequence as $ t $ ranges from 1 to $ T $ \citep{LanchantinDeepmotifdashboard2017}. Fortunately, the LSTM outputs a classification at each time step and we can simply record the output vector given by $ \mathrm{softmax}(\textbf{V}\textbf{h}_t)$ at each time step $ t $ to obtain the temporal output scores, where $\textbf{h}_t\in(-1,1)^d$ is the hidden output of the LSTM \citep{lipton_critical_2015} with $ d $ units, and $\textbf{V}\in\mathcal{R}^{C\times d} $ is the output weight matrix. Note that nothing changes during training where the single label of an input sequence is compared to the classification only at the final time step $ T $.

These predictions at each time step can then be superimposed onto the original signal, either as colour-coded predicted categories or as the probability of being in the correct class. For the latter, we record the element in the softmax vector corresponding to the true label of the input sequence. As we show later, this temporal output score technique is limited by its cumulative nature; it provides the tipping points during model decision making, but it does not indicate which input features were most salient.

\subsection{Input derivatives}\label{sec:input_deriv}
\newcommand{\point}{\textbf{x}^{(0)}}
\newcommand{\inp}{\textbf{x}}
Owing to the LSTM being highly nonlinear, it is difficult to determine the influence of each input $ \inp $ on $ s_c $. We can, however, approximate $ s_c(\inp) $ with a linear function in the neighbourhood of a specific input sequence $ \point $ by using the first-order Taylor expansion\footnote{$ s_c $ is assumed to be smooth at $ \point $.}
\begin{small}
	\begin{align*}
	s_c(\inp) &\approx \left(\nabla s_c(\inp)|_{\point}\right)^T\inp + s_c(\point) - \left(\nabla s_c(\inp)|_{\point}\right)^T\point. \label{eq:input_deriv}
	\end{align*}
\end{small}
This linear formulation allows us to see that the magnitude of the elements in $ \nabla s_c(\inp)|_{\point} $ determines the importance of the corresponding elements in $ \inp $. Visualizing these magnitudes enables interpretation of the model decisions. The gradient $ \nabla s_c(\inp)|_{\point} $ can, fortunately, be computed with a single step of backpropagation. However, in section \ref{sec:input_feature} we show that this is a bad approximation for LSTMs for reasons explained in \citet{FongInterpretableExplanationsBlack2017a}. Instead of the arbitrary variations explored with this approach, we could follow a more controlled method -- perturbing the input by deleting (occluding) subsections.

\subsection{Occlusion}\label{sec:occ}
\citet{ZeilerVisualizingUnderstandingConvolutional2014} proposed iteratively occluding regions of an image in order to visualize which regions are salient. We apply this to the sequential paradigm of LSTMs by occluding subsequences of an input sequence with a predetermined occlusion value.

This technique iteratively occludes subsequence $ \textbf{x}_{t':t'+w} $ of $ \textbf{x}_{1:T} $ by some constant $ k $ as $ t' $ ranges from 1 to $ T $, where $ w $ denotes the width of the occlusion. At the start and end of the sequence, the occlusion is shortened such that each element of $ \textbf{x}_{1:T} $ is occluded the same number of times. Over these iterations, a new sequence $ \textbf{v}_{1:T} $ accumulates the class score $ s_c $ of a sequence $ \textbf{x}_{1:T} $ in element $ \textbf{v}_t $ if element $ \textbf{x}_t $ is not occluded. If an input $ \textbf{x}_t $ is more important for classification, then its corresponding sum of class scores $ \textbf{v}_t $ will be high. For visualization, the sequence $ \textbf{v}_{1:T} $ is scaled to range from zero to one and superimposed onto the original input sequence. For images processed in scanline order, we can use more creative occlusions, such as a $ 5\times5 $ pixel block. For $ 28\times 28 $ pixel images, this tailored filter would occlude 5 contiguous pixels every 28 time steps, repeated 5 times.

The manual nature of the occlusion approach limits its efficacy. First, the correct occlusion value $ k $ that resembles a ``deletion'' of information is difficult to determine. Second, the width $ w $ and the structure of the occluded regions have to be specified and are fixed, leaving endless possible combinations unexplored. Next, we introduce a method that mitigates these issues by employing gradients to efficiently explore the space of possible input perturbations.

\subsection{Learning an input mask}\label{sec:mask}
With focus on convolutional neural network (CNN) applications, \citet{FongInterpretableExplanationsBlack2017a} recently proposed a method for learning a mask that optimally deletes information in an image. Here we describe an adaptation that makes their proposal suitable for LSTMs applied to physiological signals.

The aim is to find a mask $ \textbf{m}_{1:T}$ that minimally deletes information in the input sequence $ \textbf{x}_{1:T} $ while minimizing the class score $ s_c $. With $ d $ denoting the number of inputs per time step $ t $, we have $\textbf{m}_t\in[0,1]^d $. The function $ \phi $ that masks each element of the input is defined as
\begin{equation}
\phi(\textbf{x}_{1:T};\textbf{m}_{1:T}) = \textbf{m}_{1:T}\odot\textbf{x}_{1:T} + k(\textbf{1}-\textbf{m}_{1:T}),
\end{equation}
where $ k $ is a constant resembling a deletion of information. We seek a mask with elements that are near binary, either deleting completely or not at all, and with as few zero-elements as possible. The $ L_1 $ norm provides a suitable technique to encourage this property, but also provides enticing conditions for the mask to trigger artefacts in the LSTM \citep{FongInterpretableExplanationsBlack2017a}. To prevent artefacts, the total variation of $ \textbf{m}_{1:T} $ is added to the objective function, which encourages the mask to be smooth. The resulting objective function is given by
\begin{align*}
\obj=&\argmin_{\textbf{m}_{1:T}} \lambda_1||1-\textbf{m}_{1:T}||_1 +  \lambda_2\sum_{t=1}^{T-1}|\textbf{m}_{t+1}-\textbf{m}_t|\\ &+ s_c\big(\phi(\textbf{x}_{1:T};\textbf{m}_{1:T})\big), \numberthis \label{eq:mask_obj}
\end{align*}
where $ \lambda_1 $ and $ \lambda_2 $ weight our preference for zero-valued mask elements and mask smoothness. For visualization the learned mask is scaled to range from zero to one, subtracted from one, and superimposed onto the original input.

\citet{FongInterpretableExplanationsBlack2017a} additionally upsample the input images and jitter them to further mitigate the triggering of artefacts. Essentially, this jitter randomly translates the input by a few elements. By minimizing the expected class score over these random translations, the mask is required to have a similar effect on neighbouring elements. In our experiments, we did not find signs of artefacts and applying jitter to the input resulted in larger-than-required masks. 

All of our experiments used {\em Adam} \citep{KingmaAdamMethodStochastic2015} with a learning rate of 0.001 to optimize the mask over 500 iterations. Among initial values of 0, 0.5, and 1 for the mask, the latter convincingly yielded the best results. We experimented with various values for $ \lambda_1 $ and $ \lambda_2 $. Setting $ \lambda_2=0.001 $ provided good results over all the datasets analysed and values of $ \lambda_1$ are shown in section \ref{sec:input_feature}. Learning an input mask is insensitive to these hyperparameters; changing them changes the size of the important regions but the important features remain the same.

\section{Results}
In order to apply these visualization techniques, we require a trained LSTM. Thus two LSTM classifiers are trained, each on a different dataset, using the cross-entropy objective function. The datasets comprise the MNIST dataset \citep{lecun_mnist_1998}, processed in scanline order, and the MIT-BIH arrhythmia dataset \citep{myMoodyimpactMITBIHArrhythmia2001}. Similar to medical time series, the MNIST dataset, processed in scanline order, consists of continuous-valued sequences, which enables intuitive evaluation of the visualization techniques. 

For the MIT-BIH dataset, single heartbeats were extracted from longer filtered signals on channel 1 by means of the {\em BioSPPy} package \citep{biosppy}. The signals were filtered using a bandpass FIR filter between 3~Hz and 45~Hz. The Hamilton QRS detector \citep{HamiltonOpensourceECG2002} was used to detect and segment single heartbeats. We chose the four heartbeat classes that are best represented over different patients in the dataset: normal, right bundle branch block (RBBB), paced, and premature ventricular contraction (PVC). This resulted in 89,670 heartbeats from 47 patients. We randomly split the data over patients to have heartbeats from 33 train-, 5 validation-, and 9 test-patients (70:10:20). An acceptable split was considered to have all classes in each set contain at least $0.9\gamma \times $\textit{smallest-class-size} data points, where $ \gamma $ is the split-fraction (0.7, 0.1, or 0.2). For MNIST the standard data split was used and all the examples shown in this section are from test sets.

The LSTMs, without peepholes, are based on \citet{lipton_critical_2015} and were trained with Adam at a learning rate of 0.001, a dropout probability of 0.1, and a minibatch size of 200. Optimization was run for 100 epochs and the validation loss was used to determine the best model. With 2 layers of 128 hidden units, an accuracy of 98.7\% was achieved on MNIST, similar to \citet{arjovsky2016unitary} and \citet{cooijmans2016recurrent}. With a single layer of 128 units and 0.001 weight decay, an accuracy of 80.3\% was achieved on the MIT-BIH Arrhythmia dataset. We manually explored hyperparameter values to find adequate results.

\subsection{Qualitative evaluation}\label{sec:qual_viz}

\subsubsection{Temporal output scores}
\begin{figure}
	\hspace{-0.35cm}	
	\begin{subfigure}{.38\linewidth}
		\centering
		\includegraphics[width=\linewidth]{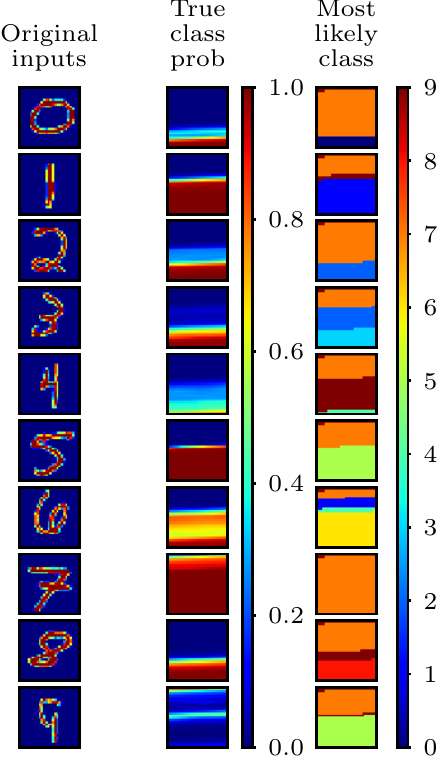}
	\end{subfigure}
	\hspace{0.25cm}
	\begin{subfigure}{.62\linewidth}
		\centering
		\includegraphics[width=\linewidth]{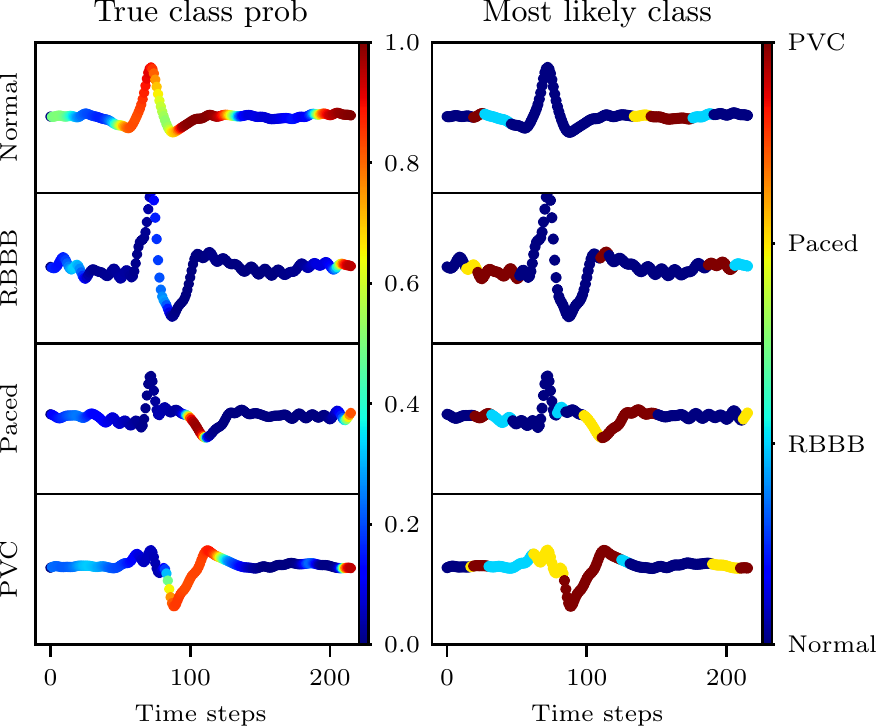}
	\end{subfigure}
	\caption[Temporal output scores]{Temporal output scores for the MNIST dataset (\textit{left}) and the MIT-BIH dataset (\textit{right}). The probability that the LSTM assigns to the true class is shown, along with the predicted most likely class. For the MIT-BIH dataset, the 4 heartbeat classes are normal, right bundle branch block (RBBB), paced, and premature ventricular contraction (PVC). The ECG signals are plotted with the same y-scale. Note, the model incorrectly classified the nine-digit example and all ECG examples were correctly classified. (Best viewed electronically.)
	}  
	\label{fig:temporal}
\end{figure}

In figure \ref{fig:temporal} we illustrate the temporal output scores for the MNIST and MIT-BIH datasets. For both datasets, the probability of being in the correct class, as well as the most probable class, is displayed at each time step.

Usefully, this technique allows tracking of LSTM decisions over time. Considering the most likely class on MNIST, the digits are all classified as nine for the first six time steps before the prediction switches to seven -- an interesting prior learned by the model. The one-digit marginally constitutes the largest class of the MNIST training and validation datasets, with seven being the close second. We speculate that the model is quicker to identify a seven (it has a long flat region at the top) than a one, which could explain the learned prior. With such balanced datasets, the prior could also depend on the order of the data points seen during the final epoch of training. On the other hand, the normal class of the MIT-BIH dataset is much larger than the other categories, which explains why the model initially assumes all ECGs to be normal.

A neat feature of this technique, not provided by any of the other visualization techniques explored, is the ability to determine the minimum sequence length required for classification. Compare, for example, classification of the five-digit, which is confidently correct from halfway through the sequence, to the classification of the four-digit, which switches to the correct class at the very last moment -- it would be detrimental to shorten the MNIST digit sequences. Similarly, for the MIT-BIH dataset, the LSTM switches to the correct class at the very last moment for all of the examples.

Although this is a fun and easy-to-implement technique, it is limited to cumulative sequential explanations. For example, in the RBBB heartbeat example, {\em what} the model looks at is not apparent. The technique depicts when the predictions are changed, but not the extent to which each time step contributed to the prediction.

\subsubsection{Input feature salience} \label{sec:input_feature}
In this section, the input derivative, occlusion, and mask learning visualization techniques (sections \ref{sec:input_deriv} to \ref{sec:mask}) are visually compared. For the deletion techniques we experimented with 0, 0.5, and the mean for values of $ k $. For the occlusion width $ w $, we explored values of 2, 5, 10, 15, 20, 25, and 50 on both datasets. Of the various hyperparameters explored for each visualization technique, we present the set of most informative results.

\begin{figure*}[t]
	\centering
	\begin{minipage}[c]{0.64\textwidth}
		\includegraphics[width=\textwidth]{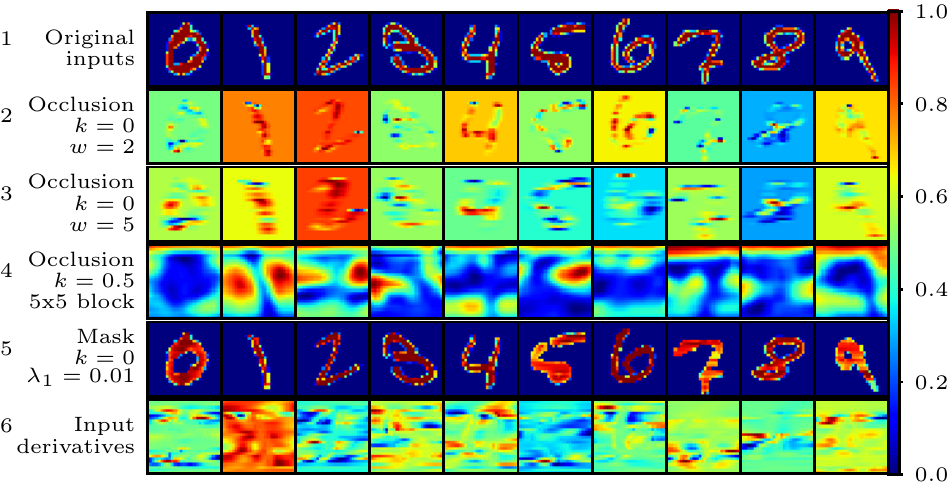}
	\end{minipage}\hfill
	\begin{minipage}[c]{0.35\textwidth}
		\caption{Input feature salience for examples of the MNIST dataset. The different features that are important according to the occlusion, mask, or input derivative techniques, are displayed on a scale of 0 to 1, where 1 is important. Each column shows the analysis of the same, correctly classified, original input. The occlusion width is denoted by $ w $ and the deletion value by $ k $. The $ 5\times 5  $ block refers to a carefully structured occlusion, which is the equivalent of a $ 5\times 5  $ pixel block being occluded for each iteration (see section \ref{sec:occ}). Most of the techniques find some interpretable salient features, with the mask learning technique, row~5, producing the best results.
		}
		\label{fig:mnist_sample_saliency}
	\end{minipage}\hfill
\end{figure*}

We start the comparison on the easy-to-understand MNIST dataset in figure \ref{fig:mnist_sample_saliency}. The standard occlusion technique successfully finds interpretable salient features, however, when the occlusion width $ w=2 $, many seemingly important features go unnoticed, for example, the seven-digit in row~2. On the other hand, setting $ w=10 $ results in overly elongated salient regions and exemplifies the difficulty of hyperparameter selection for this technique.

In the domain of MNIST digits, setting the deletion value $ k=0 $ is sensible because, given the digit strokes, the background is unimportant and zero-valued. Nevertheless, occluding with $ k=0.5 $ yields interpretability by producing the negative of the original input, and the most effective occlusion with $ k=0.5 $ is achieved when using a carefully structured $ 5\times5 $ block occlusion (row~4). When $ k=0 $, however, the $ 5\times5 $ block occlusion is less effective than the standard $ w=5 $ occlusion.

Evidently, learning a mask is the most effective salience technique for this task (row~5). Setting the deletion value $ k=0 $ results in a mask that covers most of the digit stroke patterns, which is intuitively the best option. \citet{FongInterpretableExplanationsBlack2017a} use the mean of the input as the deletion value, which in our case yields less interpretable results. It would seem that the selection of the deletion value $ k $ is task-specific.

\begin{figure}
	\centering
	\includegraphics[width=0.4\linewidth]{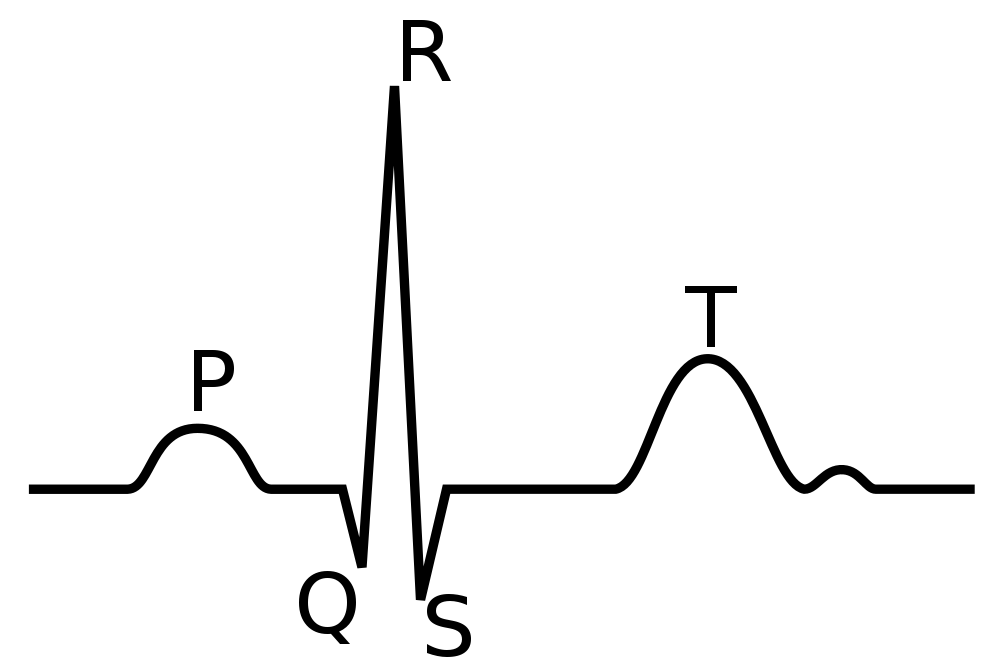}
	\caption{A standard single-lead ECG with the interest points labelled.}
	\label{fig:qrs}
\end{figure}

\begin{figure}
	\centering
	\includegraphics[width=\linewidth]{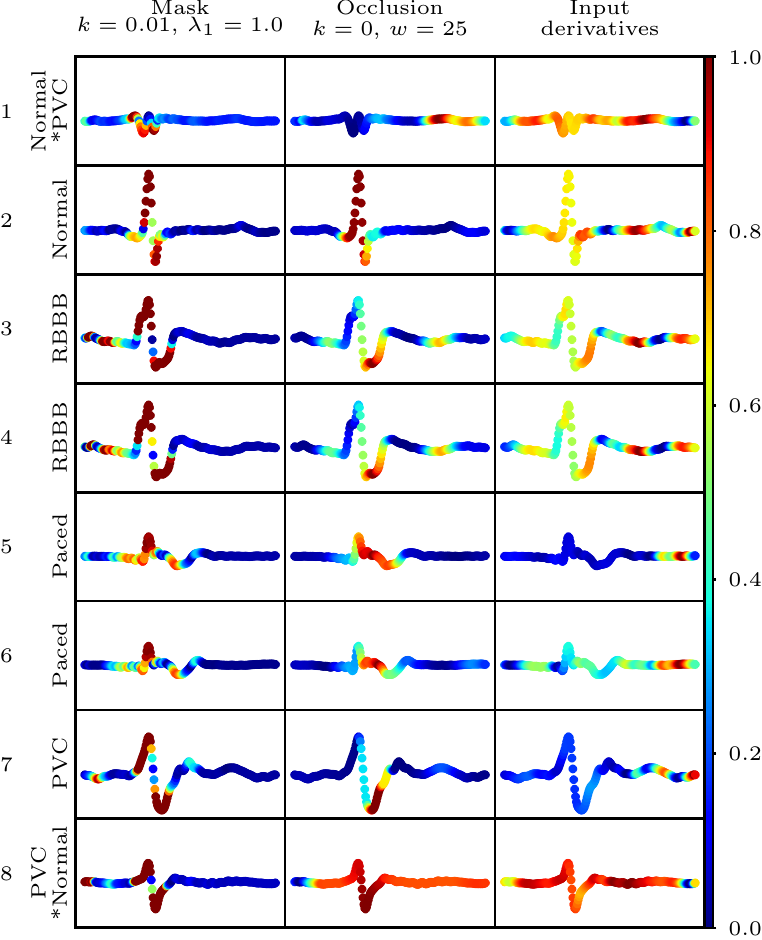}
	\caption[Input feature salience for the MIT-BIH dataset]{Input feature salience for ECG signals of the MIT-BIH dataset. The input feature importance for LSTM classification, as given by the occlusion, mask, and input derivative techniques, are displayed on a scale of 0 to 1, where 1 is important. The deletion value is denoted by $ k $, and the occlusion width by $ w $. On the left-hand side, the true label of the ECG is indicated, with the five classes of heartbeats being normal, right bundle branch block (RBBB), paced, and premature ventricular contraction (PVC). Where the LSTM incorrectly classified an input, the correct class is indicated by *. All the signals are displayed with the same y-scale and have a length of 216 time steps (x-axis).
	}
	\label{fig:ecg_sample_saliency}
\end{figure}

With a better understanding of how deletion values $ k $ and occlusion widths $ w $ influence visualizations, we proceed to visualize salient features for the LSTM on the MIT-BIH dataset. A standard ECG with annotated cardiologist interest points, P, Q, R, S, and T, is shown in figure \ref{fig:qrs}. With occasional reference to these interest points, we discuss the comparison of the different visualization techniques on the MIT-BIH dataset, as illustrated in figure \ref{fig:ecg_sample_saliency}.

At first glance, it's evident that the occlusion and mask techniques have some overlap (rows 2, 3, 4, 7, and 8). However, later in our discussion, it becomes apparent that the additional regions explained by the learned mask render it superior to the other techniques. Over different examples in each class, similar features are found salient by the mask and occlusion techniques. Furthermore, as expected in clinical practice \citep{myMoodyimpactMITBIHArrhythmia2001}, the ECG leads varied among subjects (see the difference between rows 1 and 2), which makes generalization harder for LSTMs.

We consulted a cardiologist to help identify the salient features detected by LSTMs that align with medical theory. The analysis is summarized as follows:

\begin{itemize}
	\item We begin with the normal heartbeat; the model correctly identifies the QRS complex as important, with the learned mask indicating that more importance is placed on the R-peak and S-wave (row 2). Because the signal in row 1 is from a different ECG lead, it has a low R-peak, which given the importance of the R-peak for normal heartbeats could explain the misclassification. As found in practice, the model finds the Q-wave to be less helpful.
	
	\item A wide S-wave is usually seen in right bundle branch block (RBBB) heartbeats, which was correctly identified as salient by the model -- highlighted by the mask and occlusion techniques for both RBBB examples (rows 3 and 4). The learned mask additionally shows that the LSTM correctly identifies the extra bump leading up to the R-peak (a characteristic of an RSR pattern) to be important for classifying heartbeats. Note that RBBB heartbeats can look similar to LBBB heartbeats depending on the ECG lead.
	
	\item Depending on where the pacing leads are placed within the heart, the R-wave (or sometimes Q-wave) follows the pacing spike (see rows 5 and 6). Thus identifying the narrow upstroke of the pacing spike could provide a means of detecting paced heartbeats. According to the learned mask, the LSTM learns to identify this pacing spike, whereas the occlusion technique seems to find the R-to-S transition important.
	
	\item Lastly, in rows 7 and 8, the learned mask shows that the model considers the ratio of the R-peak and S-wave as an important feature of premature ventricular contraction. Medically this is relevant for some ECG leads, with the S-wave being deep relativeto the R-peak. In practice, the duration of the QRS complex is primarily used to determine premature ventricular contraction, but it's difficult to justify whether this is something the model finds salient.
\end{itemize}

The analysis demonstrates that the input features considered salient align well with medical theory. Note, however, cardiologists usually classify heartbeat arrhythmias by means of multi-lead ECG signals and take the current patient health status (e.g., chest pain) into account. Such additional inputs could improve LSTM performance.

A visualization technique that has not been given much attention thus far is the input derivative. On both of the before mentioned datasets, this technique yields the least interpretable salient features. An additional advantage of the occlusion and mask techniques is to determine the effect of unwanted perturbations on classification, such as missing values in physiological signals.

We observed the visualization techniques on a significant portion of the datasets and found the salient regions to be consistent for each class -- the examples presented in this section represent their entire corresponding dataset. The following section describes a quantitative effort to find a measure of efficacy for the entire test dataset.

\subsection{Quantitative evaluation}
In this section, we compare the efficacy of the input feature salience techniques by calculating how much they reduce the class score $ s_c $ on average. Here the total size of the salient regions is not a concern, instead, the important input regions should be as relevant as possible regardless of their size. To compare the different techniques, however, we need to penalize larger deletion areas because simply deleting the whole input could yield the largest score reduction. Therefore, we scale the score reduction for each input sequence by the ratio $ \frac{T-M^{(n)}}{T} $, where $ M^{(n)} $ is the number of input elements occluded of the sequence $ \xn_{1:T} $, and $ T $ is the total number of time steps. Before occluding a sequence and computing the reduced class score, a threshold denoted by $ \alpha\in[0,1] $ has to be specified, above which, the input features are considered important enough to occlude.

Note, this metric does not consider whether the occlusions are correct, for example, the ECGs could be occluded at clinically irrelevant regions. It solely provides a measure of how efficiently each technique finds salient input features for the LSTM. Hence, establishing the utility of the visualization techniques still requires a qualitative evaluation. To have the average class score reductions be related to the utility of the salience technique, we select the hyperparameters based on visual analysis. We set $ k=0 $ for all techniques and use $ w=5 $ for MNIST occlusion and $ w=25 $ for MIT-BIH occlusion.

\begin{figure}[h]
	\centering
	\includegraphics[width=\linewidth]{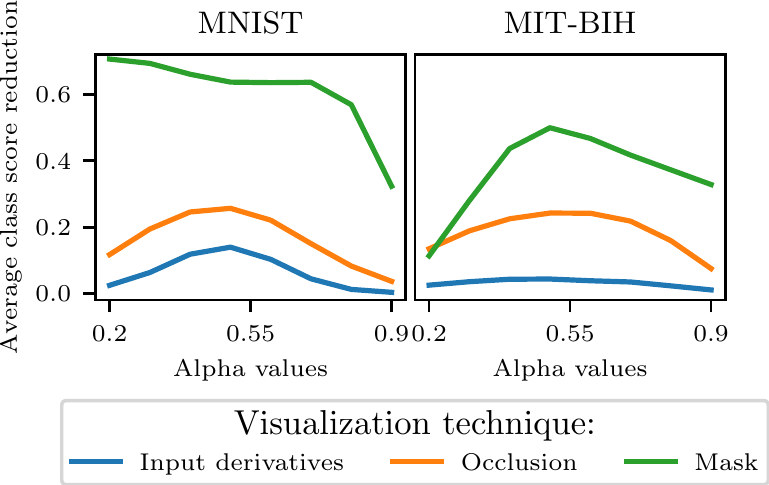}
	\caption{Average class score reductions for the three input feature salience techniques. The y-scale is the same for each graph. Learning a mask is clearly the superior technique.}
	\label{fig:quant}
\end{figure}

In figure \ref{fig:quant} we illustrate the average score reductions over a range of values for $ \alpha $. Evidently, the learned mask most efficiently deletes information from the input. In contrast to the other techniques, when $ \alpha=0.9 $, the learned mask still greatly reduces the class score, meaning that the most salient features are truly important. 

\section{Discussion and conclusion}
This paper deals with the important problem of providing insights into how black box models, specifically LSTMs, make their decisions. The premise is simple -- for many practical applications, one needs justification of the decision, instead of a bare prediction. LSTMs and other neural networks do not provide this kind of information. When these models are applied to medical data, understanding why they make certain decisions would allow clinicians to build better trust in them and could provide novel insights on medical phenomena.

This work goes some way to improving our understanding of LSTMs. We compared the efficacy of four visualization techniques for the LSTM. In this comparison, it is argued that explaining what an LSTM focusses on depends in large part on the meaning of varying the input to the model. It was found that learning a deletion mask yields the most interpretable results and the largest reduction of the class score. Performance of this technique seemed fairly unaffected by different deletion values, but this could be data-dependent. Furthermore, we found that the ECG input features considered salient by LSTM align well with medical theory. 

While performing experiments we also investigated class mode visualization \citep{SimonyanDeepConvolutionalNetworks2013}, but the results were completely uninterpretable. We have reported results for only univariate inputs. During our analysis, visualization techniques were applied to multivariate physiological signals from intensive care unit patients and traumatic brain injury patients. In such multivariate scenarios, the visualizations become convoluted and lose their interpretability. Understanding what LSTMs look for in multivariate signals thus remains an open problem.

\section*{Acknowledgements}	
We thank Vadir Baktash and Steve Foulkes for useful discussions on medical theory.

\bibliography{references/My_Library,references/references}

\begin{thebibliography}{30}
\providecommand{\natexlab}[1]{#1}
\providecommand{\url}[1]{\texttt{#1}}
\expandafter\ifx\csname urlstyle\endcsname\relax
  \providecommand{\doi}[1]{doi: #1}\else
  \providecommand{\doi}{doi: \begingroup \urlstyle{rm}\Url}\fi

\bibitem[Arjovsky et~al.(2016)Arjovsky, Shah, and Bengio]{arjovsky2016unitary}
Arjovsky, Martin, Shah, Amar, and Bengio, Yoshua.
\newblock Unitary evolution recurrent neural networks.
\newblock In \emph{International {{Conference}} on {{Machine Learning}}}, pp.\
  1120--1128, 2016.

\bibitem[Bach et~al.(2015)Bach, Binder, Montavon, Klauschen, M{\"u}ller, and
  Samek]{BachPixelWiseExplanationsNonLinear2015}
Bach, Sebastian, Binder, Alexander, Montavon, Gr{\'e}goire, Klauschen,
  Frederick, M{\"u}ller, Klaus, and Samek, Wojciech.
\newblock On {{Pixel}}-{{Wise Explanations}} for {{Non}}-{{Linear Classifier
  Decisions}} by {{Layer}}-{{Wise Relevance Propagation}}.
\newblock \emph{PloS one}, 10\penalty0 (7):\penalty0 e0130140, 2015.

\bibitem[Bahdanau et~al.(2015)Bahdanau, Cho, and
  Bengio]{BahdanauNeuralMachineTranslation2015}
Bahdanau, Dzmitry, Cho, Kyunghyun, and Bengio, Yoshua.
\newblock Neural {{Machine Translation}} by {{Jointly Learning}} to {{Align}}
  and {{Translate}}.
\newblock In \emph{International {{Conference}} on {{Learning
  Representations}}}, 2015.

\bibitem[Carreiras et~al.(2015)Carreiras, Alves, Louren\c{c}o, Canento, Silva,
  Fred, et~al.]{biosppy}
Carreiras, Carlos, Alves, Ana~Priscila, Louren\c{c}o, Andr\'{e}, Canento,
  Filipe, Silva, Hugo, Fred, Ana, et~al.
\newblock {BioSPPy}: Biosignal processing in {Python}, 2015.
\newblock URL \url{https://github.com/PIA-Group/BioSPPy/}.
\newblock (Accessed on 03/28/2018).

\bibitem[Ching et~al.(2018)Ching, Himmelstein, {Beaulieu-Jones}, Kalinin, Do,
  Way, Ferrero, Agapow, Zietz, Hoffman, Xie, Rosen, Lengerich, Israeli,
  Lanchantin, Woloszynek, Carpenter, Shrikumar, Xu, Cofer, Lavender, Turaga,
  Alexandari, Lu, Harris, DeCaprio, Qi, Kundaje, Peng, Wiley, Segler, Boca,
  Swamidass, Huang, Gitter, and Greene]{ChingOpportunitiesObstaclesDeep2018}
Ching, Travers, Himmelstein, Daniel~S., {Beaulieu-Jones}, Brett~K., Kalinin,
  Alexandr~A., Do, Brian~T., Way, Gregory~P., Ferrero, Enrico, Agapow,
  Paul-Michael, Zietz, Michael, Hoffman, Michael~M., Xie, Wei, Rosen, Gail~L.,
  Lengerich, Benjamin~J., Israeli, Johnny, Lanchantin, Jack, Woloszynek,
  Stephen, Carpenter, Anne~E., Shrikumar, Avanti, Xu, Jinbo, Cofer, Evan~M.,
  Lavender, Christopher~A., Turaga, Srinivas~C., Alexandari, Amr~M., Lu,
  Zhiyong, Harris, David~J., DeCaprio, Dave, Qi, Yanjun, Kundaje, Anshul, Peng,
  Yifan, Wiley, Laura~K., Segler, Marwin H.~S., Boca, Simina~M., Swamidass,
  S.~Joshua, Huang, Austin, Gitter, Anthony, and Greene, Casey~S.
\newblock Opportunities {{And Obstacles For Deep Learning In Biology And
  Medicine}}.
\newblock \emph{bioRxiv}, pp.\  142760, 2018.
\newblock \doi{10.1101/142760}.

\bibitem[Choi et~al.(2016{\natexlab{a}})Choi, Bahadori, Schuetz, Stewart, and
  Sun]{ChoiDoctorAIPredicting2016}
Choi, Edward, Bahadori, Mohammad~Taha, Schuetz, Andy, Stewart, Walter~F., and
  Sun, Jimeng.
\newblock Doctor {{AI}}: {{Predicting Clinical Events}} via {{Recurrent Neural
  Networks}}.
\newblock \emph{JMLR Workshop Conf Proc}, 56:\penalty0 301--318, August
  2016{\natexlab{a}}.
\newblock ISSN 1938-7288.

\bibitem[Choi et~al.(2016{\natexlab{b}})Choi, Bahadori, Sun, Kulas, Schuetz,
  and Stewart]{ChoiRETAINInterpretablePredictive2016a}
Choi, Edward, Bahadori, Mohammad~Taha, Sun, Jimeng, Kulas, Joshua, Schuetz,
  Andy, and Stewart, Walter.
\newblock {{RETAIN}}: {{An Interpretable Predictive Model}} for {{Healthcare}}
  using {{Reverse Time Attention Mechanism}}.
\newblock In \emph{Advances in {{Neural Information Processing Systems}}}, pp.\
   3504--3512. {Curran Associates, Inc.}, 2016{\natexlab{b}}.

\bibitem[Choi et~al.(2017)Choi, Bahadori, Song, Stewart, and
  Sun]{ChoiGRAMGraphbasedAttention2017}
Choi, Edward, Bahadori, Mohammad~Taha, Song, Le, Stewart, Walter~F., and Sun,
  Jimeng.
\newblock {{GRAM}}: {{Graph}}-based {{Attention Model}} for {{Healthcare
  Representation Learning}}.
\newblock In \emph{Proceedings of the 23rd {{ACM SIGKDD International
  Conference}} on {{Knowledge Discovery}} and {{Data Mining}}}, pp.\  787--795,
  New York, NY, USA, 2017. {ACM}.
\newblock ISBN 978-1-4503-4887-4.
\newblock \doi{10.1145/3097983.3098126}.

\bibitem[Clifford et~al.(2017)Clifford, Liu, Moody, Lehman, Silva, Li, Johnson,
  and Mark]{CliffordAFClassificationshort2017}
Clifford, Gari~D, Liu, Chengyu, Moody, Benjamin, Lehman, Li, Silva, Ikaro, Li,
  Qiao, Johnson, AE, and Mark, Roger~G.
\newblock {{AF Classification}} from a short single lead {{ECG}} recording: The
  {{PhysioNet}}/{{Computing}} in {{Cardiology Challenge}} 2017.
\newblock \emph{Computing in Cardiology (CinC)}, 44, 2017.

\bibitem[Cooijmans et~al.(2016)Cooijmans, Ballas, Laurent, G{\"u}l{\c{c}}ehre,
  and Courville]{cooijmans2016recurrent}
Cooijmans, Tim, Ballas, Nicolas, Laurent, C{\'e}sar, G{\"u}l{\c{c}}ehre,
  {\c{C}}a{\u{g}}lar, and Courville, Aaron.
\newblock Recurrent batch normalization.
\newblock \emph{arXiv preprint arXiv:1603.09025}, 2016.

\bibitem[Deming et~al.(2016)Deming, Targ, Sauder, Almeida, and
  Ye]{DemingGeneticArchitectDiscovering2016}
Deming, Laura, Targ, Sasha, Sauder, Nate, Almeida, Diogo, and Ye, Chun~Jimmie.
\newblock Genetic {{Architect}}: {{Discovering Genomic Structure}} with
  {{Learned Neural Architectures}}.
\newblock \emph{arXiv:1605.07156 [cs, stat]}, May 2016.

\bibitem[Fong \& Vedaldi(2017)Fong and
  Vedaldi]{FongInterpretableExplanationsBlack2017a}
Fong, Ruth~C. and Vedaldi, Andrea.
\newblock Interpretable {{Explanations}} of {{Black Boxes}} by {{Meaningful
  Perturbation}}.
\newblock In \emph{Proceedings of the {{IEEE International Conference}} on
  {{Computer Vision}}}, pp.\  3429--3437, 2017.

\bibitem[Hamilton(2002)]{HamiltonOpensourceECG2002}
Hamilton, Patrick~S.
\newblock Open source {{ECG}} analysis.
\newblock \emph{Computing in Cardiology (CinC)}, pp.\  101--104, September
  2002.
\newblock \doi{10.1109/CIC.2002.1166717}.

\bibitem[Jagannatha \& Yu(2016)Jagannatha and Yu]{jagannatha2016bidirectional}
Jagannatha, Abhyuday~N. and Yu, Hong.
\newblock Bidirectional {{RNN}} for {{Medical Event Detection}} in {{Electronic
  Health Records}}.
\newblock In \emph{Proceedings of {{NAACL}}-{{HLT}}}, pp.\  473--482, 2016.

\bibitem[Karpathy et~al.(2015)Karpathy, Johnson, and
  {Fei-Fei}]{KarpathyVisualizingUnderstandingRecurrent2015}
Karpathy, Andrej, Johnson, Justin, and {Fei-Fei}, Li.
\newblock Visualizing and {{Understanding Recurrent Networks}}.
\newblock \emph{arXiv:1506.02078 [cs]}, June 2015.

\bibitem[Kindermans et~al.(2016)Kindermans, Sch{\"u}tt, M{\"u}ller, and
  D{\"a}hne]{KindermansInvestigatinginfluencenoise2016}
Kindermans, Pieter, Sch{\"u}tt, Kristof, M{\"u}ller, Klaus, and D{\"a}hne,
  Sven.
\newblock Investigating the influence of noise and distractors on the
  interpretation of neural networks.
\newblock \emph{arXiv:1611.07270 [cs, stat]}, November 2016.

\bibitem[Kingma \& Ba(2015)Kingma and Ba]{KingmaAdamMethodStochastic2015}
Kingma, Diederik and Ba, Jimmy.
\newblock Adam: {{A Method}} for {{Stochastic Optimization}}.
\newblock In \emph{International {{Conference}} on {{Learning
  Representations}}}, 2015.

\bibitem[Lanchantin et~al.(2017)Lanchantin, Singh, Wang, and
  Qi]{LanchantinDeepmotifdashboard2017}
Lanchantin, Jack, Singh, Ritambhara, Wang, Beilun, and Qi, Yanjun.
\newblock Deep motif dashboard: Visualizing and understanding genomic sequences
  using deep neural networks.
\newblock In \emph{Pacific {{Symposium}} on {{Biocomputing}}}, pp.\  254--265.
  {World Scientific}, 2017.
\newblock ISBN 978-981-320-780-6.
\newblock \doi{10.1142/9789813207813_0025}.

\bibitem[LeCun(1998)]{lecun_mnist_1998}
LeCun, Yann~A.
\newblock The {{MNIST Database}} of {{Handwritten Digits}}.
\newblock \emph{http://yann.lecun.com/exdb/mnist/}, 1998.

\bibitem[Li et~al.(2016)Li, Chen, Hovy, and
  Jurafsky]{LiVisualizingUnderstandingNeural2016}
Li, Jiwei, Chen, Xinlei, Hovy, Eduard, and Jurafsky, Dan.
\newblock Visualizing and {{Understanding Neural Models}} in {{NLP}}.
\newblock In \emph{Proceedings of the {{Conference}} of the {{North American
  Chapter}} of the {{Association}} for {{Computational Linguistics}}: {{Human
  Language Technologies}}}, 2016.

\bibitem[Lipton et~al.(2015{\natexlab{a}})Lipton, Berkowitz, and
  Elkan]{lipton_critical_2015}
Lipton, Zachary~C., Berkowitz, John, and Elkan, Charles.
\newblock A {{Critical Review}} of {{Recurrent Neural Networks}} for {{Sequence
  Learning}}.
\newblock \emph{arXiv:1506.00019 [cs]}, May 2015{\natexlab{a}}.

\bibitem[Lipton et~al.(2015{\natexlab{b}})Lipton, Kale, Elkan, and
  Wetzell]{lipton_learning_2015}
Lipton, Zachary~C., Kale, David~C., Elkan, Charles, and Wetzell, Randall.
\newblock Learning to {{Diagnose}} with {{LSTM Recurrent Neural Networks}}.
\newblock \emph{arXiv:1511.03677 [cs]}, November 2015{\natexlab{b}}.

\bibitem[Mahendran \& Vedaldi(2016)Mahendran and
  Vedaldi]{MahendranSalientDeconvolutionalNetworks2016}
Mahendran, Aravindh and Vedaldi, Andrea.
\newblock Salient {{Deconvolutional Networks}}.
\newblock In \emph{Computer {{Vision}} \textendash{} {{ECCV}}}, pp.\  120--135.
  {Springer, Cham}, October 2016.
\newblock ISBN 978-3-319-46465-7 978-3-319-46466-4.
\newblock \doi{10.1007/978-3-319-46466-4_8}.

\bibitem[Moody \& Mark(2001)Moody and Mark]{myMoodyimpactMITBIHArrhythmia2001}
Moody, George~B. and Mark, Roger~G.
\newblock The impact of the {{MIT}}-{{BIH Arrhythmia Database}}.
\newblock \emph{IEEE Engineering in Medicine and Biology Magazine}, 20\penalty0
  (3):\penalty0 45--50, May 2001.
\newblock ISSN 0739-5175.
\newblock \doi{10.1109/51.932724}.

\bibitem[Rajkomar et~al.(2018)Rajkomar, Oren, Chen, Dai, Hajaj, Liu, Liu, Sun,
  Sundberg, Yee, Zhang, Duggan, Flores, Hardt, Irvine, Le, Litsch, Marcus,
  Mossin, Tansuwan, Wang, Wexler, Wilson, Ludwig, Volchenboum, Chou, Pearson,
  Madabushi, Shah, Butte, Howell, Cui, Corrado, and
  Dean]{RajkomarScalableaccuratedeep2018}
Rajkomar, Alvin, Oren, Eyal, Chen, Kai, Dai, Andrew~M., Hajaj, Nissan, Liu,
  Peter~J., Liu, Xiaobing, Sun, Mimi, Sundberg, Patrik, Yee, Hector, Zhang,
  Kun, Duggan, Gavin~E., Flores, Gerardo, Hardt, Michaela, Irvine, Jamie, Le,
  Quoc, Litsch, Kurt, Marcus, Jake, Mossin, Alexander, Tansuwan, Justin, Wang,
  De, Wexler, James, Wilson, Jimbo, Ludwig, Dana, Volchenboum, Samuel~L., Chou,
  Katherine, Pearson, Michael, Madabushi, Srinivasan, Shah, Nigam~H., Butte,
  Atul~J., Howell, Michael, Cui, Claire, Corrado, Greg, and Dean, Jeff.
\newblock Scalable and accurate deep learning for electronic health records.
\newblock \emph{arXiv:1801.07860 [cs]}, January 2018.

\bibitem[Simonyan et~al.(2013)Simonyan, Vedaldi, and
  Zisserman]{SimonyanDeepConvolutionalNetworks2013}
Simonyan, Karen, Vedaldi, Andrea, and Zisserman, Andrew.
\newblock Deep {{Inside Convolutional Networks}}: {{Visualising Image
  Classification Models}} and {{Saliency Maps}}.
\newblock \emph{arXiv:1312.6034 [cs]}, December 2013.

\bibitem[Springenberg et~al.(2014)Springenberg, Dosovitskiy, Brox, and
  Riedmiller]{SpringenbergStrivingSimplicityAll2014}
Springenberg, Jost~Tobias, Dosovitskiy, Alexey, Brox, Thomas, and Riedmiller,
  Martin.
\newblock Striving for {{Simplicity}}: {{The All Convolutional Net}}.
\newblock \emph{arXiv:1412.6806 [cs]}, December 2014.

\bibitem[Teijeiro et~al.(2017)Teijeiro, Garc{\'\i}a, Castro, and
  F{\'e}lix]{TeijeiroArrhythmiaClassificationAbductive2017}
Teijeiro, Tom{\'a}s, Garc{\'\i}a, Constantino~A., Castro, Daniel, and
  F{\'e}lix, Paulo.
\newblock Arrhythmia {{Classification}} from the {{Abductive Interpretation}}
  of {{Short Single}}-{{Lead ECG Records}}.
\newblock \emph{arXiv:1711.03892 [cs]}, November 2017.

\bibitem[Xu et~al.(2015)Xu, Ba, Kiros, Cho, Courville, Salakhudinov, Zemel, and
  Bengio]{XuShowAttendTell2015b}
Xu, Kelvin, Ba, Jimmy, Kiros, Ryan, Cho, Kyunghyun, Courville, Aaron,
  Salakhudinov, Ruslan, Zemel, Rich, and Bengio, Yoshua.
\newblock Show, {{Attend}} and {{Tell}}: {{Neural Image Caption Generation}}
  with {{Visual Attention}}.
\newblock In \emph{International {{Conference}} on {{Machine Learning}}}, pp.\
  2048--2057, June 2015.

\bibitem[Zeiler \& Fergus(2014)Zeiler and
  Fergus]{ZeilerVisualizingUnderstandingConvolutional2014}
Zeiler, Matthew~D. and Fergus, Rob.
\newblock Visualizing and {{Understanding Convolutional Networks}}.
\newblock In \emph{Computer {{Vision}} \textendash{} {{ECCV}}}, pp.\  818--833.
  {Springer, Cham}, September 2014.
\newblock \doi{10.1007/978-3-319-10590-1_53}.

\end{thebibliography}
\bibliographystyle{icml2018}

\end{document}